\def\year{2022}\relax
\documentclass[letterpaper]{article} 
\usepackage{aaai22}  
\usepackage{times}  
\usepackage{helvet}  
\usepackage{courier}  
\usepackage[hyphens]{url}  
\usepackage{graphicx} 
\usepackage{natbib}  
\usepackage{caption} 
\DeclareCaptionStyle{ruled}{labelfont=normalfont,labelsep=colon,strut=off} 
\usepackage{algorithm}
\usepackage{algorithmic}

\usepackage{newfloat}
\usepackage{listings}
\floatstyle{ruled}
\newfloat{listing}{tb}{lst}{}
\floatname{listing}{Listing}
\copyrighttext{Presented at the AI-HRI Symposium at AAAI Fall Symposium Series (FSS) 2022}
\title{Human Autonomy as a Design Principle for Socially Assistive Robots}
\author {
    Jason R. Wilson
}
\affiliations{
    Franklin \& Marshall College\\
    Lancaster, PA 17603\\
    jrw@fandm.edu
}

\begin{document}

\maketitle

\begin{abstract}
High levels of robot autonomy are a common goal, but there is a significant risk that
the greater the autonomy of the robot the lesser the autonomy of the human working with the robot.  For vulnerable populations like older adults who already have a diminished level of autonomy, this is an even greater concern.
We propose that human autonomy needs to be at the center of the design for socially assistive robots.  Towards this goal, 
we define autonomy and then provide architectural requirements for social robots to support the user's autonomy.
As an example of a design effort, we describe some of the features of our \textit{Assist} architecture.
\end{abstract}

As robots become more popular and are more regularly tasked with helping individuals accomplish day-to-day tasks, it is vital that these robots respect the human state of the people they are helping. For some robots, this need is more critical, as they are assisting individuals in vulnerable populations.
There is a particular need to improve the designs of socially assistive robots (SARs), which are designed to assist people through social interaction \cite{feil2005defining} and are often employed to help individuals with disabilities, older adults, and children \cite{martinez2020socially, papadopoulos2020systematic}. 
These populations are already at greater risk of having their autonomy inhibited or diminished \cite{nordenfelt2004}. 
A disabled person has restricted autonomy as a result of the constraints on their abilities. 
Older adults, often challenged by illness and disability, may struggle with a loss of autonomy as their skills and abilities diminish.
Children are still developing the physical and mental skills necessary to autonomously make decisions that align with productive goals.
With these populations already having their autonomy at risk, introducing an assistive robot into the person's care system adds further risk that the person assisted by the robot will have their autonomy negatively impacted.

To better understand autonomy and the effects a robot may have on it, we review perspectives from the psychological, medical, and human-robot interaction (HRI) literature. 
From psychology, self-determination theory argues that autonomy, 
along with competence and relatedness, are the essential elements that together form a basic framework of motivation \cite{deci1991motivation}.
Many types of autonomy are discussed in the medical and healthcare literature, particularly in relation to the physician-patient relationship, where the freedom and ability to decide on a treatment plan and be able to follow it are necessary elements of a patient's autonomy \cite{Valero2019}.
In the field of HRI, autonomy is a relatively new topic. It has mostly been discussed in the context of designing mechanisms for a robot to moderate the amount of assistance it provides \cite{Greczek2015,wilson2020challenges}.

Significantly more work is needed in HRI, where human autonomy needs to be a design principle in the development of assistive robots to prevent impeding on vulnerable populations' already diminished autonomy. In this work, we outline four concepts describing autonomy and provide examples of how these concepts relate to HRI design.
For robots autonomously deciding how it assists,
there are some necessary architectural features to help support the user's autonomy.
We outline three architectural requirements related to a robot's social skills.
As an example of one approach working towards meeting these requirements, we provide a short description of our new \textit{Assist} architecture. 
It is designed to support the autonomy of the user by adapting its assistance to the user's needs and providing transparency in how its reasoning aligns with the goals and preferences of the person it is helping.

\section{Characteristics of Autonomy}

In order to understand how to design robots to better support user autonomy, it is necessary to identify a standard definition of autonomy for its use in HRI. 
\textit{We define human autonomy as someone's ability to act independently and freely make decisions regarding themselves, including decisions that align with their values and are not a product of coercion or outside pressure}. 
This definition reflects four overlapping concepts describing many of the characteristics of a design that are important for supporting autonomy: \textit{independence, choice, control,} and \textit{identity}. 
\textit{Independence} means having the ability and liberty to do some task or action. 
Having freedom of \textit{choice} means making rational decisions regarding one's own physical or cognitive condition, which align with ones self-interest, and carrying out those decisions over time.
\textit{Control} is the degree to which the individual is able to express control over their environment.
\textit{Identity} is a person's narrative, goals, beliefs, desires, and other personal qualities that encompass who they are and what they value and are often the basis for the decisions that they make.

\subsection{Human Autonomy in HRI}

There are some examples of how characteristics of each of these dimensions have been applied in the design of assistive robots.  Some have focused on adapting the design process to be inclusive of the target population, thereby allowing topics of autonomy to arise during the design of the robot and to allow the development of the robot to consider autonomy from its earliest stages of development.  For example, designers of a robot to guide visually impaired persons included designers with visual impairments \cite{Azenkot2016}.  As a result, they identified during the design process many design features related to autonomy, such as providing \textit{choices} regarding when the robot comes to help them, whether to schedule assistance, and how the robot will guide or aid them.
Other research has focused on developing hardware and software components that allow for the support of autonomy in user interactions.
This can include supporting a person's \textit{independence} by designing  physical components to periodically cleaning objects off of the floor, thus removing trip hazards for older adults or people with disabilities \cite{Fischinger2016}.  For social robots, the \textit{identity} of the person can be fostered using software components enabling the robot to recognize a person's individuality, whether it be preferences regarding private spaces or addressing a person by their preferred name. 
Lastly, for a person to maintain \textit{control} over their environment when the robot is assisting, there needs to be communication.  A social robot may respond to a person's verbal expressions, as well as some of their nonverbal \cite{wilson2020challenges, wilson2021enabling}.  A robot that can explain its behavior also creates an opportunity for the person to exert further control by providing feedback to the robot on its reasoning so that the robot can adapt its future behavior \cite{wilson2020knowledge}.
Going forward, improving the communication capabilities of the robot, enabling it to better perceive, understand, and use social signals, will equip the robot with the tools to better support a person's autonomy.

\section{Architectural Requirements}


Human caregivers rely on a variety of social skills to help regulate the dynamic system of interaction. 
They modify their behavior in response to what they perceive to be the person's needs, receive feedback, and may take further action to help the person.  Much of what happens may rely on implicit cues and other forms of nonverbal communication.  The caregiver may recognize the person is stuck because of a gesture or facial expression the person makes, and then gives a suggestion for how to proceed.  To see if that got the person unstuck the caregiver may watch the person for further cues and to help decide whether more help is needed.  If the person looks confused by the suggestion, and the caregiver may explain why it made that suggestion.

Socially assistive robots similarly need to be able to respond to cues and determine how best to help.
We propose the following capabilities are critical for a socially assistive robot to protect the autonomy of the user:
\begin{itemize} \itemsep0em
    \item Recognize a person's needs, goals, and preferences.
    \item Adapt help to the individual.
    \item Communicate intent and justify its actions.
\end{itemize}

Recognizing a person's needs can include inferences specific to the task they are trying to accomplish.  The robot can observe their task and predict what action the user needs to take next or what assistance would help the person complete the task.  However, purely task-based reasoning is unlikely to protect the individual's autonomy.  Constantly providing perfect assistance outlining every correct step the user needs to take leaves the user with few choices.  

One of the challenges is recognizing when to help and how much help is needed.
One approach is to wait for help to be explicitly requested.  However, this could also lead to waiting too long to help. 
People may be shy or hesitant to ask for help, or unsure if they can or should ask for help.  
As a result, a robot needs to be able to recognize and interpret both explicit and implicit social cues used to communicate how much help a person needs and wants.
Implicit cues may include expressions of frustration or gazing regularly at the robot to get confirmation.  When a person is looking for confirmation, not a lot of help is expected.  Whereas if the person appears frustrated, more help may be needed.


Adapting the robot's assistance to the particular person and situation is another challenge.  
For a social robot to support a person's autonomy, it must be able to find a balance between too little and too much help.  Too much help can cause a person to feel controlled and pressured to make choices that they do not understand or are inconsistent with the goals. 
Too little help can also inhibit a person's independence. Insufficient help may leave the person without the necessary resources and thus inhibiting the person from completing the task on their own.
In addition to adjusting to the particular amount of help that is needed, the robot should also adapt to the user's goals and preferences.  A person may prefer to complete a task in a particular manner or have a desired outcome in mind.  
As a robot decides how best to help, the choices need to be consistent with the user's expectations.  Suggestions by the robot to proceed differently may be viewed as attempts to control the person or simply not respecting the person's individuality.

The reasons for a robot's actions, how it helps and why, may not always be clear.  
Blindly following the robot's help can leave a person feeling controlled.  This feeling may be magnified if the user feels the robot is helping too much or suspects that the robot is making a mistake.
The robot's reasoning may be flawed and in need of correction.  After a robot explains its reasoning, 
the user can identify where the reasoning went wrong.
As a result, the user takes more control of the situation by helping the robot not make the same mistake again.








\section{Assist Architecture}


We are designing the \textit{Assist} architecture to enable a social robot to provide assistance.  
The architecture provides the robot with some capabilities to recognize needs, adapt its assistance, and explain its reasoning.
We describe here how the capabilities afforded by this architecture are intended to support the autonomy of the user.
We do not suggest that our solution is complete.
Our intent in describing these features is to provide an example of how to approach designing capabilities to protect a user's autonomy.

\vspace{-2mm}
\subsection{Recognizing Needs}

When an assistive robot helps, there is a risk that the robot takes too much control of the situation.  For example, with the robot to aid visually impaired persons, designers rejected the idea that the robot approach an individual that it perceives needs guidance \cite{Azenkot2016}.  Instead, they preferred to be able to decide if and when they want assistance, thereby remaining in control of their situation and environment.  Similarly in the Assist architecture, the robot provides an opportunity for the person to act on their own and uses social cues exhibited by the person to indicate that the person wants some amount of help.

To determine when the robot should help, the architecture recognizes some verbal cues and eye gaze patterns \cite{wilson2021enabling}.  The robot captures the spoken audio, which is put through an automatic speech recognition tool to get the text of the user's speech.  The robot then uses a shallow processing of the text to determine if the user's speech indicates that help is being requested or is needed.
Simultaneously, the robot captures video of the user's face to monitor eye gaze patterns.  In particular, it attempts to recognize two gaze patterns that indicate that the user may be requesting help: mutual gaze and confirmatory gaze \cite{kurylo2019using}.  When the user is initiating mutual gaze, they redirect their gaze from the task they are performing to the robot.  If the robot is speaking, this change in gaze simply is attending to the robot's speech.  However, if the robot is not speaking, this gaze change can indicate that the user is attempting to get the robot's attention.  The robot may be expected to check the work, provide confirmation of correctness, hint at what is wrong, or ask if the person would like help.  The confirmatory gaze pattern is when a person directs their gaze back and forth between the robot and the task.  This behavior is used to direct the robot's attention to a recently completed action in order to receive feedback.

Combining perceptions of the user's speech and gaze, we determine how much help the user needs.  Our scale of needing help has five levels:
\begin{enumerate}
    \item Flow: the user is in a state of flow and should not be interrupted.
    \item Hesitation: the user's flow may have been disrupted, but no help may be needed yet.
    \item Confirmation: the user is attempting to check the correctness of their actions.
    \item Inquiry - Self: the user is seeking information but may be attempting to find the information on their own.
    \item Inquiry - Other: the user realizes the information may best be obtained from another agent (i.e., the robot).
\end{enumerate}

\vspace{-2mm}
\subsection{Adaptive Assistance}

After recognizing the user's needs, the robot decides how best to help.  In the robot's decision-making process, the user's autonomy is one of the principle concerns.  To promote the user's independence, the robot needs to determine the appropriate amount of assistance and provide only enough help to enable the user to complete the task as independently as possible.

We use the Hint Engine \cite{wilson2018needbased} that plans out the remaining steps in a task, where the first step in that plan is the basis for the robot's assistance.  It selects a hint or suggestion for that step that also corresponds to the amount of help the person needs.  
For each action in a task, we model four levels of assistance that can be provided for that action.  The four levels, derived from the PASS Manual \cite{Holm1997} used in occupational therapy, describe assistance that is \emph{verbal supportive}, \emph{verbal non-directive}, \emph{verbal directive}, and \emph{gestures}.  Each level of assistance provides more help, by  being more direct or providing more information.  The intent is that by matching the level of help needed to the level of help provided, the robot can help maximize the person's sense of autonomy \cite{wilson2018towards}.

Along with the social cues suggesting how much help the user needs, the robot's decisions take into account the preferences of the user.  Preferences are represented as constraints in the robot's reasoning and alters the plans formed by the Hint Engine \cite{wilson2020knowledge}.  
If the preference alters the first step in the plan, then the robot will choose a different means of helping because the first step in the plan is the basis for the robot deciding how to help.

\vspace{-2mm}
\subsection{Explanation}

A robot explaining the reasoning behind its help and suggestions can 
provide information to aid a person in their future choices.  Additionally, the robot can assure the person that the robot is not trying to control the person by describing how its assistance relates to the goals of a particular task and is consistent with the preferences of the person.
In the Assist architecture, we construct causal explanations of the robot's assistance, alerting the user to potential risks while also recognizing an individual's preferences \cite{wilson2020knowledge}.  
Future work will use these explanations as a mechanism for receiving user feedback, allowing the robot to identify flaws or misconceptions in its reasoning and make corresponding updates.

\section{Discussion}

Human autonomy needs to be a design principle for socially assistive robots.  For many years, researchers have indirectly worked towards this goal.  Often it is pursued while focusing on effective collaboration by adapting robot behavior \cite{rossi2017user}.  For human autonomy to be an explicit goal of the design, we first need to define autonomy.  Our definition reflects many perspectives on autonomy and provides a starting point.  However, we expect this definition will need further revision.  The four characteristics of autonomy we describe (independence, control, choice, and identity) are intentionally overlapping concepts, but sometimes the distinctions can become vague.  When a person makes a choice, are they exercising their independence, exerting control over the situation, or expressing their identity?  While it can be all three, it may be important to identify which concept is most central, and clearer distinctions between the concepts would aid in this identification.

Given this definition, we also propose that for a software architecture of a socially assistive robot to support human autonomy, it must be able to recognize a person's needs, adapt its assistance to the individual, and explain its reasoning.   
These requirements may not be complete, and further work may reveal additional requirements.  
Additionally, we suggest that while all of the requirements are essential for a complete approach to supporting autonomy, a robot may still provide some support if lacking in any of these features.  For example, a robot unable to explain itself 
can protect a person's autonomy by recognizing and responding to their needs, but the long-term effectiveness may be hindered by the user not being able to understand the robot's behavior and thereby altering it.

A major gaps that remains is how to determine whether the design is successful at supporting the user's autonomy.
An essential step would be designing metrics to measure the impact on the user's autonomy.
Additionally, we speculate that a robot may be able to indirectly self-evaluate through feedback from the user and regular adaptation.  A user that wants their autonomy maintained may give the robot feedback, implicitly working towards this goal.




\section*{Conclusion}

A robot needs many social skills in order to support the autonomy of a user.  Our architecture works towards this goal and addresses the four concepts of autonomy we describe.  
In particular, it addresses a person's \textit{independence} and \textit{control} by regulating how much the robot assists in response to the social cues the robot perceives.
By recognizing and adapting to a user's preferences, it also supports a person's \textit{identity}.
As the explanation components develop further and allow the person to provide feedback to the robot, the person will be provided with more \textit{choices} in how the robot functions.




The concepts of independence, control, choice, and identity form the basis for some design principles for socially assistive robots.
By focusing on autonomic principles in the robot design and implementation, one can develop robots which not only support user autonomy but are also perceived as more helpful and more ethical. 

\bibliography{sample}

\end{document}